# Online Sequential Extreme Learning Machines: Features Combined From Hundreds of Midlayers

Chandra Swarathesh Addanki
Dept of Computer Science
caddanki@lakeheadu.ca
Lakehead University

*Abstract* — **In this paper, we develop an algorithm called hierarchal online sequential learning algorithm (H-OS-ELM) for single feed feedforward network with features combined from hundreds of midlayers, the algorithm can learn chunk by chunk with fixed or varying block size, we believe that the diverse selectivity of neurons in top layers which consists of encoded distributed information produced by the other neurons offers better computational advantage over inference accuracy. Thus this paper proposes a Hierarchical model framework combined with Online-Sequential learning algorithm, Firstly the model consists of subspace feature extractor which consists of subnetwork neuron, using the sub-features which is result of the feature extractor in first layer of the hierarchy we get rid of irrelevant factors which are of no use for the learning and iterate this process so that to recast the the subfeatures into the hierarchical model to be processed into more acceptable cognition. Secondly by using OS-Elm we are using non-iterative style for learning we are implementing a network which is wider and shallow which plays a important role in generalizing the overall performance which in turn boosts up the learning speed .**

## I. INTRODUCTION

With the rapid evolution and success in the field of deep learning we have seen the rise of various models such as Feed-Forward Neural Network, Long Short Term Memory , Generative Adversarial Network(GAN),Convolutional Neural Networks, Recurrent Neural Networks (RNNs), Conditional Random Fields (CRFs). In fact the development of Deep Learning has started way back in 1980's , In the recent times we have seen various Deep Learning models such as AlexNet , AmoebaNet , RNN (Recurrent neural network ) ,these ,models have advanced object recognition by discovering the hidden structures in high dimensional data, the only restrictions with these models is that they are Batch learning and long training time due to these particular reason the training time is quite slower, furthermore because of the Back propaganda used in this kind of iterative learning the model suffers from the false local Minima and vanishing gradient problems and the change in learning rate has direct consequences with the accuracy, there are various kinds of Non-iterative learning models which can solve these problems such as SVM (Support Vector Machines) , Random Forest

But these learning strategies are rarely explored. Furthermore with the development of DL models more layers are added to the hierarchy, but with the increase in the layers the computational cost is also increased and it will take long time to train the model , making it very harder during inference stage so it motivates us to design an hierarchical model which is computational efficient and can provide us with performance that is not much less than the model which learns features from end to end training.

In this paper we introduce the following deep three layered network.

- Feature Extractor which is the first general layer in our model, it is Subnetwork node which consists of hundreds of two layered network which is a function or constructed by other nodes , The feature extractor is used for generating subspace features from the input data.

- Then in the second general layer the optimal features extracted are combined from the local extractor which help in distinguishing the classes, As the features from the same class have identical features whereas the features from different classes have different features. thus, optimal solution can be obtained from combining.

- The third general layer is used for classification.

## II. Proposed Method

Online Sequential Extreme Learning Machines is one of the important emerging machine learning techniques. The main aspects of these techniques is that they do not need a learning process like deep learning or machine learning to calculate the parameters of the models.

Consider an Input data X which has **m** samples and **N** dimensions where '**w**' is the input weight, '**b**' is the bias and θ represents activation function, **β** represents the beta value which is obtained by using **Moore's Penrose pseudoinverse** on hidden nodes and a single-layered feed forward network **f** with **d** neurons can be represented mathematically as

$$\sum_{j=1}^{L} \beta_j \phi(\mathbf{w}_j \mathbf{x}_i + b_j), \ i \in [1, N].$$

(1)

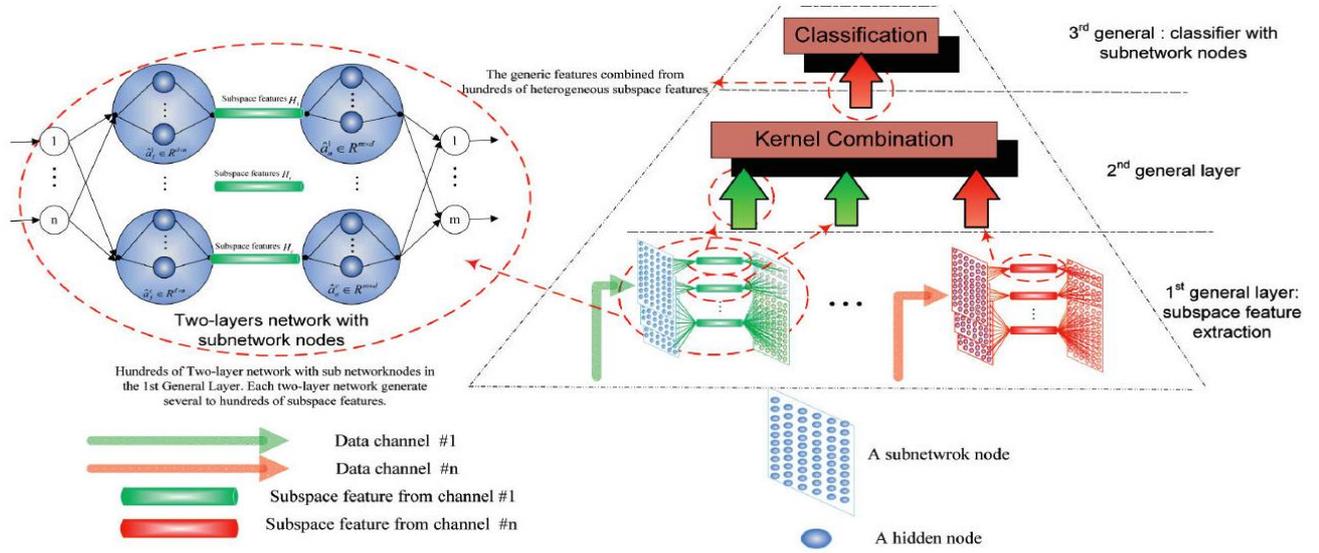

Fig. 1. The proposed method with multiple data channel structure.

The theory behind the Elm States that weight of the hidden layers does not need the tuning and thus it is independent of the training data.

the following equation represents the relation between train label , the inputs and output of the network are

$$y_i = \sum_{j=1}^{L} \beta_j \phi(w_j x_i + b_j) = t_i + \epsilon_i, \; i \in [\![1, N]\!],$$

(2)

The transformations are done in hidden nodes using the input data into a different interpretation in two steps:

- Firstly, the input data is projected into the hidden layer through the weights and biases of the input layer.
- applying non-linear activation function to the result of first step.

ELMs are mathematically solved as common Neural Networks in their matrix form. The matrix form is represented here:

$$H = \begin{bmatrix} \phi(w_1 x_1 + b_1) & \cdots & \phi(w_L x_1 + b_L) \\ \vdots & \ddots & \vdots \\ \phi(w_1 x_N + b_1) & \cdots & \phi(w_L x_N + b_L) \end{bmatrix},$$

$$\beta = \left(\beta_1^T \cdots \beta_L^T\right)^T, \; T = \left(y_1^T \cdots y_N^T\right)^T.$$

(3)

Given 'T' which is the target we want to reach, a unique solution with least squared error can be found using **Moore-Penrose generalized inverse**.

Therefore, using one single compute operation the values of the weights of the hierarchical nodes that will result in the solution with the least error to predict the target **T** can be solved using. .

$$H\beta = T$$
$$\beta = H^\dagger T$$

(4)

Unlike the multilayer network structure, in this paper, we propose a new multilayer Hierarchal network architecture

the proposed architecture has two main differences when compared with other network architecture:

- In this paper, it is assumed that neurons itself can be composed of other neurons
- In each node we are generating beta value from its associated feature
- The proposed architecture does not follow the conventional rule in traditional architecture that is all the nodes should be connected, in our proposed architecture the subnetwork nodes are not connected.
- In the proposed architecture, multiple data streams are used in the First layer, the deep features extracted shows diverse and heterogeneous characteristics.

**B. First General Layer: Subspace Feature Extraction**

**Step 1**: Given a dataset with M data samples where $\{x_k\}$ is the feature and $\{y_k\}$ is the target such **that** $\{x_k, y_k\}_{k=1}^{m}$, $x^k \in R^n$ from a continuous system and subnetwork node index to '0 '

**Step 2**: Generate a new subnetwork node randomly and generate beta using *Morres-Penrose Pseudoinverse,* where $a^f \in R^{d \times n}$, $b^f \in R$ is the random subnetwork neuron. $H_c^f$ is the cth subspace features

$$C = c+1$$
$$H_c^f = a_c^f . x + b_c \quad (5)$$

Step 3 : Given **target 'y'**, the subnetwork nodes in the Firstlayer , the *Morre-Penrose pseudoinverse* and the beta is calculated using the equation (4)

$$a_h^c = ((H_c^f)^T * H_c^f) *y \quad (6)$$

$$b_h^c = \sqrt{mse((a_h * H_f^c cT) - (y))} \quad (7)$$

Step 4: residual error e is calculated

$$e = y - (a_h^c * H_f^c + b_h^c) \quad (8)$$

Step 5: The error feedback data is calculated
$$\quad (9)$$
$$P = e . (a_h^h)^{-1}$$

Step 6: Update the normalized error feedback data as

$$P = u (P + H_f^c) \quad (10)$$

Step 7: update the parameters of subnetwork node in first layer as

$$a_{temp} = P . (xx^t)^{-1} \quad (11)$$

$$a_f^c = a_f^c + \lambda(a_{temp} - a_f^c) \quad (12)$$

$$b_f^c = \sqrt{mse(a_f^c . x - p}, b_f^c \in R \quad (13)$$

Step 8 : generate $c^{th}$ subspace feature
$$H_c^f = a_c^f . x + b_c \quad (14)$$

Generate L-1 subnetwork nodes using steps 1-8 and obtain the L subspace features

**D. Second General Layer: Subspace Feature Combination**

The features extracted from the second layer are combined in this layer , to combine the features, combination operator is used

$$H^{I \oplus J} = K(H_f^i, H_f^j, \gamma)$$
$$\quad (15)$$

**E. Third General Layer : Classification with subnetwork nodes**

Given M distinct feature samples which is generated from the combination operator (H,t) , a sigmoid function 'g' and a normalization function u(x0 -> (0,1] and for continuous output we have

$$\lim_{n \to +\alpha} \| t - ((g(a_p^1, b_p^1, H)). B_p^1 + .. + -((g(a_p^c, b_p^c, H)). B_p^c \| = 0$$, which consists of the probability
Calculate the subnode $(a_p^c, b_p^c)$

$$a_p^c = g^{-1}((e_{c-1} . H^T(HH^T))^{-1} \quad (16)$$

$$b_p^c = \frac{sum(a_p^c . H - h^{-1}((e_{c-1})))}{N}, b_p^c \in R \quad (17)$$

$$g^{-1}(.) = -\log(1/(.) - 1) \quad (18)$$

$$e_c = t - u_n^{-1} . g(H, a_p^c, b_p^c) \quad (19)$$

$$B_p^c = (e_{c-1}, u^{-1}(g . (a_p^c . H + b_p^c))) / (g . a_p^c . H + b_p^c \|^2$$
$$where \ [.]^{-1} \ is \ the \ inverse \ function \quad (21)$$

## II . Algorithm

### A. Proposed Algorithm

Given N features groups$(Q_1, \ldots, Q_n)$ generated from the same data set $Q_1 = (x_k^1, y_k^1)_{k=1}^M, x_k^1 \in R^{n_1}, \ldots, N = (x_k^N, y_k^N)_{k=1}^M, x_k^N \in R^{n_N}$, positive coefficient C, and the number of sub network nodes L.

First General Layer: Subspace feature extraction: Set n=1 , and let x,y equals $Q_1$.. While $n < N$ do While $c < L$ do Obtain a subspace feature $H_f^{c+((n-1) \times L)}$ bases on steps 1-8 and generate $\beta$ using Moore's Penrose pseudoinverse. end while return obtained L subspace features $H_f^{c+(1+(n-1) \times L)} \ldots \ldots H_f^{c+((n-1) \times L)}$ from data group $Q_n$ end while return obtain $L*N$ subspace features $H_f^1 \ldots \ldots H_f^(LN)$

Second General Layer: Feature combination: Obtained combined features H as
$$H = H^{1 \oplus 2 \oplus (N \times L)} \quad (28)$$

Third General Layer: Pattern Learning
Given concatenated features H, Set c=1, e=t
While $c < L$ do
Step 1: Calculate the cth subnetwotk node $(a_p^c, b_p^c)$, and $\beta_p^c$ as

$$a_p^c = g^{-1}((e_{c-1})). x^T \left(\frac{C}{I} + xx^T\right)^{-1}$$

$$b_p^c = sum(a_p^c . H - h^{-1}((e_{c-1})))/ N, b_p^c \in R$$

$$\beta_p^c = \frac{\{e_{c-1}, u^{-1}(g(a_p^c . H + b_p^c))\}}{\|u^{-1}(g(a_p^c . H + b_p^c))\|^2}$$

OS-Elm combined with the Hierarchical network

- Training samples are divided and taken in chunk by chunk or one by one with definite or varying chunk lengths and fed into the algorithm.
- Only newly observed chunks are learned and the chunks are discarded after they are learned

In first layer features are extracted and Morres-Penrose pseudoinverse is applied in second layer the features are combined after the features are combined , finally we use the third layer to classify

## III. Experimental Verification

### A. Environment settings:

All Images are processed into grey scale, the experiment are done in repeatedly with different training and testing datasets, and the average of per-class recognition rates is recorded for each execution.

### B. Scene 15

The dataset consist of 4486 gray-value images of which 3860 images are from the 15-category scenes we use the default experimental settings, 100 images per category as training data are randomly selected and use the rest as test data. The Plus operator is us ed as the combination operator part., the number of hidden nodes is set to 100 in each sub-network node. The experimental method uses single features with a three layer network.

### C. Caltech 101

The dataset contains 9144 images in 102 object categories with the number of images per category varies from 31 to 800. Following the common experimental settings, we train on 15 and 30 samples per category and test on the rest.

### D. Cifar 10

The data set collection comprises 60000 color images in ten classes k, The number of images per class is 6000. There are 50000 pictures (5000 for each class) for training and the rest of the parts for testing.

## III Experimental results

In the experiment , we evaluate the model using different kind of dataset from small to large datasets following are the results of evaluation

TABLE I
SPECIFICATIONS OF DATASET AND EXPERIMENTAL SETTINGS

| Index | Dataset | Feature Type | Category | Subnetwork |
|---|---|---|---|---|
| 1 | Scene15[a] | Spatial Pyramid | 15 | 3 |
| 2 | Caltech101[a] | Spatial Pyramid | 102 | 3 |
| 3 | Caltech101[a] | HMP and Spatial Pyramid | 102 | 5 |
| 4 | Flower102[a] | HMP | 103 | 4 |
| 4 | Cifar10[a] | Imagenet Pretrained Deep Feature | 10 | 3 |

[a]

VI. RESULTS

| Index | Method | Dataset | Accuracy | Ti |
|---|---|---|---|---|
| 1 | Hierarchical Method[a] | Scene15 | 97.80 | |
| 2 | OS-ELM Combined Method[a] | Scene15 | 87.9 | |
| 3 | Hierarchical Method[a] | Caltech101 | 49.70 | |
| 4 | OS-ELM Combined Method[a] | Caltech101 | 50.20 | |
| 5 | Hierarchical Method[a] | Caltech101(HMP+Spatial) | 50.20 | |
| 6 | OS-ELM Combined Method[a] | Caltech101(HMP+Spatial) | 50.20 | |

[a]

## IV Conclusion

In this paper we have combined OS-elm with Hierarchical network and its performance is evaluated, the OS-elm is combined with the boosting phase of the model in the first layer and using the Moore's-penrose pseudo inverse the beta is calculated which is used for the classification in the third layer comparing our model with other state of art models it gives out better performance , we have combined sequential learning that is passing data chunks by chunks as we can see in the experimental results that even through the combined network did not give the results better than the original hierarchical network, it gave a decent performance.